\documentclass[12pt]{iopart}
\usepackage[utf8]{inputenc}
\usepackage[english]{babel}
\usepackage{balance}
\usepackage{graphics} % for pdf, bitmapped graphics files
\usepackage{epsfig} % for postscript graphics files
\usepackage{multicol}
\usepackage{subcaption}
\usepackage{balance}
\usepackage[bookmarks=true]{hyperref}
\usepackage[usenames]{color}
\usepackage{comment}
\usepackage{float}	
\usepackage{multirow}
\usepackage{multicol}
\usepackage{natbib}
\usepackage{pdflscape}
\usepackage{tabularx}
\usepackage{colortbl}
\usepackage[dvipsnames]{xcolor}
\usepackage{afterpage}

\begin{document}
\vspace*{-3.0cm}
%======================================================================
%\title{Synthetic Intelligence as Computation}
\title{Intelligence as Computation}
%======================================================================

\author{Oliver Brock}

\address{Robotics and Biology Laboratory\\Science of Intelligence\\Technische Universit\"at Berlin}
%\ead{not@needed.okay}
\vspace{10pt}

{\small
  \tableofcontents
  }

\clearpage

%======================================================================
%\title{Synthetic Intelligence as Computation}
\title{Intelligence as Computation}
%======================================================================

\author{Oliver Brock}

\address{Robotics and Biology Laboratory\\Science of Intelligence\\Technische Universit\"at Berlin}
%\ead{not@needed.okay}
\vspace{10pt}

%======================================================================
\begin{abstract}
%======================================================================
This paper proposes a specific conceptualization of intelligence as computation. This conceptualization is intended to provide a unified view for all disciplines of intelligence research. Already, it unifies several conceptualizations currently under investigation, including physical, neural, embodied, morphological, and mechanical intelligences. To achieve this, the proposed conceptualization explains the differences among existing views by different computational paradigms, such as digital, analog, mechanical, or morphological computation. Viewing intelligence as a composition of computations from different paradigms, the challenges posed by previous conceptualizations are resolved. Intelligence is hypothesized as a multi-paradigmatic computation relying on specific computational principles. These principles distinguish intelligence from other, non-intelligent computations. The proposed conceptualization implies a multi-disciplinary research agenda that is intended to lead to unified science of intelligence.
\end{abstract}

%\maketitle

%======================================================================
\section{Introduction}
%======================================================================

``The concept and notion of computation and intelligence has historically seen significant evolution which continues to this day.''~\citep{hughes_embodied_2022}

\vspace{0.5cm}

\noindent
This is not a good sign. How can we study, understand, and reproduce \textit{intelligence} from a \textit{computational} perspective, if we cannot base such efforts upon an understanding of what these terms actually mean? The lack of clarity manifests itself in the literature via a proliferation of different kinds of ``intelligences:'' computational intelligence, embodied intelligence, physical intelligence, neural intelligence, morphological intelligence, mechanical intelligence, and many more. Similar diversity of interpretation exists for the term computation, which seems inextricably linked to the notions of intelligence and cognition~\citep{chalmers_computational_2011}.

The objective of this paper is first to disentangle the notions of intelligence and computation and then to propose a unified view of intelligence (albeit not a definition!) based on these disentangled notions. The distinct emphasis of the discussion will be on the study of \textit{synthetic intelligence}, thus on the synthetic disciplines of intelligence research, including robotics, computer science, and engineering. And in the discussion that entails, I will be fully committed to a computational view of intelligence and cognition~\citep{chalmers_computational_2011}. This currently is the dominant view and, as we will see in Section~\ref{sec: computation}, I will assume a very general conceptualization of computation so that the arguments are not constrained by philosophical disputes revolving around this term.

The term \textit{synthetic intelligence} subsumes humans' attempts to re-create biological intelligence in synthetic artifacts. I avoid the term \textit{artificial intelligence}, because it is so over-used that it has become too ambiguous for a scientific discussion. Something is \textit{synthetic} if it results ``from synthesis rather than occurring naturally'' (Merriam Webster Dictionary). Therefore, it seems useful to juxtapose \textit{synthetic} with \textit{natural}, rather than with ``biological,'' as we might discover intelligence that is natural (not synthesized) but not biological.

My hope is that the conceptual clarifications of the terms \textit{intelligence} and \textit{computation} proposed in this paper can help prevent a ``Tower of Babel'' within synthetic intelligence research. Reaching consensus on the meaning of underlying concepts and terms would certainly contribute to an integrated and multi-disciplinary research agenda towards understanding the type of intelligence we observe in biological agents.

%======================================================================
\section{Argumentative Roadmap}
%======================================================================

Dualism, or more specifically \textit{mind-body dualism}~\citep{sep-dualism}, is a historical view that has influenced philosophy of mind and historical views on intelligence research. Dualism is the belief that mental processes are non-physical and thus distinct from the body. This view has a long philosophical tradition, reaching back to Plato, Aristotle, and Descartes. In the context of this paper, however, it is not necessary to go into detail on dualism, since currently the alternative view of materialism dominates science. Materialism is the belief that all processes, including those going on in the mind, are realized by material interactions, i.e., are physics-based. 

I mention dualism only because the observation that some ``shadow'' of dualism appears to remain present in intelligence research serves as a starting point for the arguments laid out in this article.  This shadow manifests itself in the distinction between brain and body, which is frequently made when talking about intelligence (see Section~\ref{sec: intelligences}). The notion of embodiment rests on the assumption that the body contributes something unique to intelligence, something distinct from what the brain contributes~\citep{varela_embodied_1993}. I will argue that this is in some sense true: the body does contribute something distinct to intelligence. But I will argue that the true nature of this unique contribution is misunderstood. The contribution does not follow from a brain/body distinction, it can be explained more compellingly.

To prepare this hopefully more compelling explanation, Section~\ref{sec: intelligences} discusses existing conceptualizations of intelligence, in particular in the light of the brain/body distinction. I will argue that this distinction is not the right one, and also not helpful, given that the brain obviously is part of the body. Hopefully, by the end of Section~\ref{sec: intelligences}, there will be the need for a new, better explanation.

The main proposal of this paper is a specific computational view of intelligence. In order to prepare describing this view, a second preparatory step will be laid out in Section~\ref{sec: computation}. This section describes different computational paradigms including digital computation, analog computation, and quantum computation.  Each of these paradigms has, in principle, the same computational power as all the others. But some problems can practically be solved much more easily in one paradigm than in another. Section~\ref{sec: computation} will present these different paradigms and attempt to differentiate them based on their ``special powers.''

After having pointed out in Section~\ref{sec: intelligences} that current conceptualizations of synthetic intelligence are problematic, possibly because they have been influenced by remnants of mind-body dualism, and after having described different types of computation in Section~\ref{sec: computation}, I will present \textit{Intelligence as Computation} in Section~\ref{sec: intelligence as computation}. In short, I will argue that intelligence is computation---so far you will probably agree. I will further argue that intelligence refers to a subset of all possible computations that implements and critically relies on yet-to-be-determined computational principles (I will propose some).

This view of intelligence as computation offers some advantages to be detailed below. But in brief: The proposed view shifts the focus from describing the generated intelligent behavior (which is the focus of most existing conceptualizations of intelligence~\citep{CambridgeHandbookOfIntelligence-2011}), onto the computational mechanisms that generate intelligent behavior. This is in some sense a return to the roots of artificial intelligence, which had viewed intelligence as symbol manipulation. This focus on computational mechanism promises to be more palatable than an attempt to characterize behavior. While behavior can vary without limit with environmental conditions, the mechanism that produces behavior is limited to the physical realization of the acting agent. A second advantage of the computational view of intelligence is, of course, that the inconsistent brain/body distinction is removed in a productive manner. Furthermore, the realization that different computational paradigms contribute differently to intelligence points to a trajectory for research and a shared conceptualization of intelligence suitable for many disciplines.

%======================================================================
\section{Intelligences---Status Quo}
\label{sec: intelligences}
%======================================================================

I will first survey some of the types of intelligences introduced in the literature, criticizing them when necessary, sometimes for being influenced by a ``dualistic shadow.'' This section also analyzes the problem that arises from these diverse, ambiguous, and sometimes inconsistent definitions. 

%----------------------------------------------------------------------
\subsection{Biological Intelligence}
%----------------------------------------------------------------------

The conceptual consensus I am aiming for in the context of synthetic research, has remained elusive in the study of \textit{biological intelligence}.  A succinct mechanistic definition and understanding of the term \textit{intelligence} does not exist~\citep{CambridgeHandbookOfIntelligence-2011}. Consensus is limited to the circular statement that ``intelligence is what an intelligence test measures.'' There is, however, relatively broad consensus that \textit{general intelligence} (which should be synonymous with \textit{intelligence}) can be subdivided into fluid (ability to process new information, learn, and solve problems) and crystallized (accumulated knowledge) intelligence~\citep{cattell_theory_1963}. A more detailed subdivision of intelligence is postulated in the theory of multiple intelligences~\citep{Gardner-93}. This theory is scientifically disputed and lacks empirical support. The list of different intelligences proposed in it includes musical intelligence, visual-spatial intelligence, linguistic intelligence, logical-mathematical intelligence, body-kinesthetic intelligence, and several more.

The proliferation of intelligences and the ongoing debate over them, together with the lack of a consensual definition of the term \textit{intelligence} itself, make it seem plausible that we have not gained an understanding sufficiently deep for conceptualizing biological intelligence. Maybe a synthetic perspective on intelligence can help make some progress.

%----------------------------------------------------------------------
\subsection{Synthetic Intelligence}
\label{sec: syntehtic intelligence}
%----------------------------------------------------------------------

I will now discuss the proliferation of intelligences within the synthetic disciplines. First, I will present definitions of various types of intelligences from the literature. I will demonstrate that some of these definitions are unduly influenced by what I called ``shadow of dualism'' earlier. I will also show that the definitions do not produce a consistent picture. To put it bluntly: when viewed together, they do not make sense. These problems, which I hope to overcome in later sections, have contributed to a vague and inflationary use of relevant concepts. Just to give one example: Some authors use the terms ``embodied'' and ``intelligence'' in the title of their publication but these words do not occur a single time within the text of the article~\citep{wang_co-design_2022}. Some people in our communities have begun to use terms such as intelligence as marketing gimmicks.

%----------------------------------------
\subsubsection{Computational Intelligence:}
%----------------------------------------

``Computational Intelligence (CI) is the theory, design, application and development of biologically and linguistically motivated computational paradigms. Traditionally the three main pillars of CI have been Neural Networks, Fuzzy Systems and Evolutionary Computation''~\citep{IEEE-CI-definition}. 
CI~\citep{Engelbrecht-2007} is a biologically inspired subset of Artificial Intelligence (the classical discipline, not the buzz word)~\citep{Russell-2020}. Conforming to some dualistic tendencies, CI is associated with the brain as opposed to the body~\citep{sitti_physical_2021}.

%----------------------------------------
\subsubsection{Physical Intelligence:}
%----------------------------------------

Physical Intelligence (PI) refers to ``physically encoding sensing, actuation, control, memory, logic, computation, adaptation, learning and decision-making into the body of an agent''~\citep{sitti_physical_2021}.  Here, the body is proposed as an alternative substrate for all functions we might associate with intelligence. It offers the view that the computations required to produce intelligence can be implemented in the brain (CI) or the body (PI). This view also seems to exhibit remnants of the dualist tradition. We might take issue with the adjective ``physical'', as both body and brain are physical entities. The adjective thus does not distinguish correctly between alternative substrates for implementing the functionality of intelligence. Even CPUs are physical. The term seems to be based on the view that digital or neural computations are not physical.

The fact that ``everything is physical'' implies that the location of computation, brain or body, might not be the most useful distinction for computational processes associated with intelligent behavior. Later in the text, I will attempt to replace distinctions based on the ``location'' of computational processes with one about \textit{how} computation is performed.

%----------------------------------------
\subsubsection{Neural Intelligence:}
%----------------------------------------

Neural Intelligence (NI) refers to all intelligent systems that rely on neurons to perform computations. These neurons can either be biological or part of an artificial neural network. In this regard, NI can be seen as a subcategory of CI. Exhibiting a ``dualistic shadow,'' the view associated with NI differentiates between body and mind: ``Biological organisms can survive and operate autonomously in unstructured environments because of the neural intelligence in their brains and the physical intelligence encoded in their bodies''~\citep{wang_physical_2022}. The idea is to transfer this dichotomy of neural and physical intelligences to ``the development of intelligent bio-inspired robots''~\citep{wang_physical_2022}. The notion of NI therefore combines a ``dualistic shadow'' with the idea of bio-mimeticism, i.e., applying insights from biological intelligence to building synthetic intelligence.

%``The intelligence of the biological agents is enabled by their neural intelligence in their brains, in the meantime, their physical intelligence encoded in the bodies plays a nonnegligible role''~\citep{wang_physical_2022}.

%----------------------------------------
\subsubsection{Embodied Intelligence:}
%----------------------------------------

``Embodied intelligence [(EI)] is the computational approach to the design and understanding of intelligent behavior in embodied and situated agents through the consideration of the strict coupling between the agent and its environment (situatedness), mediated by the constraints of the agent's own body, perceptual and motor system, and brain (embodiment)''~\citep{cangelosi_embodied_2015}. The link of brain and body finds consensus in the literature with other authors defining Embodied Intelligence (EI) as ``investigating tight coupling between an agent body and brain''~\citep{sitti_physical_2021} or ``intelligence that requires and leverages a physical body''~\citep{mengaldo_concise_2022}.

But can there be an intelligence that does not require and leverage a physical body? Certainly not, every form of intelligence has to be physically realized, irrespective of whether it is a thought realized as changes in a biological neural networks or a computation on a CPU realized as changes on finely etched circuitry. In a practical sense, nothing can be disembodied. The term \textit{Embodied Intelligence} therefore does not lead to a meaningful categorization of lines of intelligence research. It is only meaningful from the perspective of a ``dualistic shadow''in which the brain is not considered to be embodied. 

At the same time, the emphasis on coupling brain and body, which is also part of the definition of EI, is an important effort in overcoming the ``dualistic shadow.''  Also, the focus on situatedness seems crucial. For the first time, intelligence is not considered to be a phenomenon that resides inside an agent's body (considering the brain part of the body) but instead as something that arises from interactions of the agent with its environment~\citep{Pfeifer-2006,clark_supersizing_2010,Murphy-Paul-2022}. EI implies a view of computation that attributes computational capabilities to both the brain and the body.

\textit{Embodied Cognition}, in close relationship to EI, ``rejects or reformulates the computational commitments of cognitive science [to the brain], emphasizing the significance of an agent's physical body in cognitive abilities''~\citep{sep-embodied-cognition}. Certainly, what must be meant here cannot refer to a body/no-body distinction but must mean something else. Both body and brain are able to perform computations, programmed into a physical system.

%----------------------------------------
\subsubsection{Embodied Artificial Intelligence:}
%----------------------------------------

Embodied Artificial Intelligence (EAI) ``can be defined as the study of intelligent agents that can see (or more generally perceive their environment through vision, audition, or other senses), talk (i.e. hold a natural language dialog grounded in the environment), listen (i.e. understand and react to audio input anywhere in a scene.), act (i.e. navigate their environment and interact with it to accomplish goals), and reason (i.e. consider the long-term consequences of their actions)''~\citep{deitke2022retrospectives}.

Similarly to EI, there is an emphasis on the interactions between agent and environment. In contrast to the types of intelligences discussed so far, the definition of EAI~\citep{deitke2022retrospectives} does not offer anything beyond the well-established. For example, the authors of the definition attempt to distinguish EAI from robotics by statements such as ``Embodied AI also \textit{includes} work that focuses on exploring the properties of intelligence in realistic environments while abstracting some of the details of low-level control. [...] Conversely, robotics \textit{includes} work that focuses directly on the aspects of the real world, such as low-level control, real-time response, or sensor processing [emphasis added].'' But these statements about what the fields \textit{include} does not establish a distinction. The authors claim that ``there are differences in focus which make embodied AI a research area in its own right.'' But a change in focus in itself clearly does not constitute a new research area. Please note that the other notions of intelligence discussed here do not make claims of establishing a new research field.

%----------------------------------------
\subsubsection{Morphological Intelligence:}
\label{morph comp}
%----------------------------------------

The term \textit{morphology} refers to the shape and material composition of an object. Morphological Intelligence (MI) is the ``ability of a morphology to facilitate the learning of novel tasks''~\citep{gupta_embodied_2021}. This is accomplished by morphological computation (MC), which is described as ``outsourcing part of control, sensing and computation tasks to the robot body to be handled through emergence of functional behaviours due to properties such as architectured kinematics, compliance, natural resonance, damping and friction''~\citep{ghazi-zahedi_editorial_2021}.

Presumably influenced by a ``dualistic shadow,'' MI and MC juxtapose what the body does to what the brain does (Computational Intelligence, control, or computation). Implicit in the first quote is also the idea that MI is in service of learning or CI. But every form of computation is the result of the morphology of the object performing the computation. The way circuits are arranged on a VLSI chip constitute the chip's morphology and enable certain computations. Morphology cannot be a meaningful dividing line between different computational phenomena pertaining to intelligence.

%----------------------------------------
\subsubsection{Mechanical Intelligence:}
%----------------------------------------

Mechanical Intelligence (MechI) ``explores nature-inspired mechanisms for automatic adaptability and applies them to the design of structural components''~\citep{khaheshi_mechanical_2022}. Closely related to the notions of \textit{Physical Intelligence} and \textit{Morphological Intelligence}, this view of MechI emphasizes the transfer for mechanisms from nature to the production of engineering artifacts. It thus seems narrower and is subsumed by these concepts.

It is important to distinguish the above definition of MechI from Mechanical Engineering (MechE) as a field. MechE, as a research field, focuses on building functionality into physical artifacts in clever and robust ways. There are thousands of years of progress accumulated in Mechanical Engineering, and many of the views expressed above (Physical Intelligence, Embodied Intelligence, Morphological Intelligence, Mechanical Intelligence) might be viewed as subsets of Mechanical Engineering. This indicates that Mechanical Engineering can serve as an important disciplinary foundation for the study of intelligent agents and their interactions with the environment.

%----------------------------------------------------------------------
\subsection{Problems with Intelligences and Their Consequences}
%----------------------------------------------------------------------

Let us look at the consequences that arise from the definitions of different intelligences when viewed all together.

%----------------------------------------
\subsubsection{Remnants of Dualism?}
%----------------------------------------

Apparently, dualism still has an influence on the conceptual foundations of intelligence. Many of the intelligences above have a dualistic flavor. Of course, the discussion about brain versus mind and physical states versus mental states, as in Descartian dualism, is no longer present. But it seems to have been replaced by an assumed functional divide between brain and body. This is exemplified in statements like this one: ``While PI [physical intelligence] only focuses on the physical intelligence encoded in the body, EI focuses on the tight coupling between an agent's body and brain (and the environment)''~\citep{sitti_physical_2021}. This implies a relevant distinction between intelligence encoded in the body and the brain, calling the former ``physical''. \citet[Figure 1]{sitti_physical_2021} includes in physical intelligence all things we might also attribute to the brain, e.g., memory, learning, decision making, etc. The proclaimed need to coordinate brain and body in the area of Embodied Intelligence, further demonstrates this misunderstanding.

I do not see any justification for this new variant of dualism. Brain and body are close collaborators in all we might consider intelligence or cognition. There are few functions we can entirely attribute to one or the other. Perception, for example, significantly begins with task-specific pre-processing already in the retina~\citep{gollisch_eye_2010}, which is an extension of the central nervous system. Processing obviously continues in the brain, but is critically enabled by the pre-processing of information outside of the brain. Similar pre-processing happens in other senses, including touch~\citep{pruszynski_edge-orientation_2014}, hearing, and smelling~\citep{yoshioka-2013}.

Below, I will argue that we cannot divide cognitive functions into those performed by the brain and those performed by the body. It is very likely that nearly all relevant functions are implemented in a close collaboration between brain and body. Of course, I am not arguing that the brain is not a physiologically separate system. And if one wants to, one can define functions exclusively of the brain. However, if we are talking about functions that can be considered meaningful with respect to the generation of intelligent behavior, such as learning, motor control, perception, reasoning, etc., we probably should not just look to the brain or the body alone. Even memory, which one might attribute to the brain exclusively, can only work because relevant information has been filtered and structured by other parts of the body. It would be hopeless to simply store all perceptual information directly.

Maybe the new form of dualism is the consequence of an intuitive misunderstanding. Intuitively, brain and body are different in some way. However, it is not that they have sharply divided roles in terms of cognitive functions. What might justify the brain/body distinction instead is the consideration of computational paradigms. Brain and body collaborate in performing cognitive functions, but each contributes a distinct form of computation. Details about this argument will follow below, after we have discussed in Section~\ref{sec: computation} the different forms of computation we might consider here.

%----------------------------------------
\subsubsection{Bio-Inspiration}
%----------------------------------------

Many types of intelligences include bio-inspiration in their definition. This is understandable as close observation of the miracles performed by biological agents makes it blatantly obvious that our focus on brain-bound supremacy is missing an important part of the picture. This focus has been termed ``cognitive chauvinism: we tend to favor control complexity over morphological complexity. Classical artificial intelligence dispensed with the body altogether''~\citep{cangelosi_embodied_2015}. Cognitive chauvinism is probably connected to the new form of dualism discussed above.

Taking inspiration from biology seems like a very fruitful idea, and this has been proven with many inventions in everyday life. However, we should take lessons from biology without being constrained to the specific implementations of the solutions. In other words: Take the lesson, but be open to solutions that biology has not invented yet. This would mean that we should not remove bio-inspiration from our research labs and projects, as it can be incredibly productive, but we should also not simply mimic what we find.

%----------------------------------------
\subsubsection{Ambiguity}
%----------------------------------------

Many of the types of intelligences discussed above assume a dualistic stance. I argued that this stance cannot be upheld because none of the definitions identified a valid criterion for the differentiation of brain and body. Once we strip away the new dualism, Physical Intelligence becomes equivalent to intelligence and subsumes all other forms of intelligence. This is because \textit{all} relevant phenomena of all types of intelligence are physics-based, simply because there are no processes not based on physics.

The different definitions are also ambiguous and overlap in meaning. For example, Physical Intelligence, Embodied Intelligence, Morphological Intelligence, Mechanical Intelligence, and Embodied Cognition all emphasize the contribution of the body (without the nervous system). This is an indication that different intelligences are not mutually exclusive and therefore not ideal to structure intelligence research. It also stresses again the importance of finally overcoming dualism in all of its forms.

Some research programs associated with different definitions of intelligence include close ties to other types of intelligence. For example, Embodied Intelligence researchers propose \textit{mechanics} as a foundational modeling tool~\citep{mengaldo_concise_2022}, indicating that Embodied Intelligence and Mechanical Intelligence (or at least Mechanical Engineering) have so much in common that a categorical distinction may not be justified. 

\citet[Figure 3]{hadi_sadati_embodied_2022} list key aspects of Embodied Intelligence: morphological computation, co-design, material, physical intelligence, sensor design, and learning. This has significant overlap with the definition Physical Intelligence (``physically encoding sensing, actuation, control, memory, logic, computation, adaptation, learning and decision-making into the body of an agent''~\citep{sitti_physical_2021}).

The ambiguity reflected in these few examples illustrates how a more concise conceptualization of intelligence might help advance the fields by enabling a clearer conceptualization of the relevant factors.

%----------------------------------------------------------------------
\subsection{Intelligence Without Dualism}
\label{subsec: without dualism}
%----------------------------------------------------------------------

Remnants of dualism remain a recurring theme in the intelligences discussed above. Nevertheless, there is broad agreement that both body and brain---sticking to the new dualistic distinction---both contribute to the behavior of intelligent agents and must be viewed together rather than separately. 

The brain performs computation, few would dispute this. Then we are left with looking for a description of the body's contribution. Concepts like Morphological Computation, i.e.~``outsourcing part of control, sensing and computation tasks to the robot body''~\citep{ghazi-zahedi_editorial_2021}, are a response to this need and also refer to the concept of \textit{computation}. Unfortunately, the term ``morphology'' in \textit{Morphological Computation} is relatively ambiguous~\citep{EB-morphology}, but let us assume that it refers to how materials of varying properties are spatially arranged to constitute an object. This definition holds both at the macro level (objects) and at the micro level (atoms and molecules).

If you agree with this view of morphology, we are left with a dualistic dilemma: If we attribute the ability to control, sense, and compute to the body's morphology, as is suggested by the definition of MC, then we must also concede that a)~the body performs computation and b)~the arithmetic logic unit on a CPU performs its computation as a result of its morphology as well. A CPU is nothing else but different materials spatially arranged so as to realize a particular set of computations. This means that brain and body both perform computation and that this computation is always the result of morphology.

If we give up dualism (as we should) and realize that everything is physics-based, i.e., if we subscribe to materialism, all behavior arises from morphology and how it structures physical processes. There is no fundamental difference between the morphology of a CPU, a clockwork, or a robotic hand. It is all ``just'' physics, forced into a certain game plan by morphology. As a result, the distinctions among different intelligences described in this section fall apart.

We now have leveled the playing field for brains and bodies as well as for CPUs and brains. They all perform computation by structuring physics-based processes with the morphologies of their bodies---and now the body clearly includes the brain for which the exact same statement holds.

%======================================================================
\section{Computation}
\label{sec: computation}
%======================================================================

In the previous section, our discussion shifted from intelligence to computation as a way to explain how intelligent behavior comes about in a non-dualist setting. It is time to talk about computation in more principled manner, as it serves as the foundational concept for reasoning about intelligence. Of course, by computation we now mean something much broader than symbol manipulation. But how exactly should we view computation for it to serve as a useful foundation?

%----------------------------------------------------------------------
\subsection{Computation---Status Quo}
%----------------------------------------------------------------------

For most researchers in this area, computation serves as a foundation for the study of cognition~\citep{chalmers_computational_2011} and embodied cognition~\citep{sep-embodied-cognition}. I will discuss the relationship between cognition and intelligence in more detail in Section~\ref{sec: cognition versus intelligence}, but for now I will treat these two terms as synonyms, for practical reasons, and therefore assume that computationalism is also an adequate foundation for intelligence.

There are many unanswered questions about the relationship between computation and cognition. \citet{chalmers_computational_2011} lists a few of them: ``What is it for a physical system to implement a computation? Is computation sufficient for thought? What is the role of computation in a theory of cognition? What is the relation between different sorts of computational theory, such as connectionism and symbolic computation?'' These questions continue to be the subject of debate.

There are equally many questions about computation itself. As the context for our discussion is the synthesis of intelligence in physical systems, we will focus our discussion of computation in physical systems~\citep{sep-computation-physicalsystems}, called \textit{concrete} computation, as opposed to \textit{abstract} computation, addressing mathematical formalisms of computation. In the realm of concrete computation, people wonder, for example, what distinguishes a computer (believed to be a physical system capable of performing computation) from a rock (believed to not be able to perform computation). The relationship of concrete computation to abstract computation is also not fully clarified. Several theories about this relationship exist, including the simple mapping account, the semantic account, the syntactic account, the mechanistic account, and several more~\citep{sep-computation-physicalsystems}.

\textit{Pancomputationalism}~\citep{piccinini_physical_2015} is a relatively popular view of computation~\citep{sep-computation-physicalsystems}. It states that all physical systems perform computation. Ultimately, the entire universe can be seen as a giant computer, computing away at its program (the laws of physics) with morphology representing the current program state (the distribution and dynamic state of matter in space).

%----------------------------------------------------------------------
\subsection{View of Computation Used Here}
%----------------------------------------------------------------------

Pancomputationalism is helpful if we want to give up dualism. According to this view, the neurons inside the brain perform computation and so does the compliant actuation system of the body, for example, when implicitly computing a good foot posture while walking on uneven ground. Pancomputationalism offers us a view of computation that does not require a distinction between different body parts' contributions to intelligence. All physical processes are computations and since everything is physics, everything is also computation.

For the purpose of our discussion, we should, however, differentiate between useful computation and everything else. In building an intelligent agent, the computation performed by gas clouds in a far-away galaxy is of no concern to us. We can safely assume, based on the known laws of physics, that the gas clouds do not affect the behavior of our agent.

Out of all the pancomputational processes unfolding in the universe at any point in time, we call only those \textit{computations} that have been deliberately included to contribute to the behavior of the agent under consideration. In other words: we consider only those pancomputational (or physical) processes that are \textit{useful} to the agent. The outcome of the computation, including its failure, can affect the behavior of the agent. By specifying a purpose for computation---for example, the processes contributing to a person passing the driver's test---we can refer to the set of relevant physical processes as computation. We focus our notion of computation on those physical processes that affect the agent. This distinction is common in the literature on computation~\cite{Horsman-2013,Muller-2017}.

%----------------------------------------------------------------------
\subsection{Computational Paradigms}
%----------------------------------------------------------------------

We have left dualism behind and have leveled the playing field in regards to computation. Every physical thing, no matter whether it is called body or brain, performs computation, as long as it affects the behavior of the agent under consideration. To achieve this, we had to sacrifice the distinctions among different types of intelligences found in the literature~\ref{sec: intelligences}. We also had to adopt a ``practical'' definition of teleological computation so as to be able to focus on the set of computations (physical process) relevant to a particular agent.

Let us now consider this entire set of relevant computations for a particular agent. Is there a useful way to categorize them? Such a categorization could serve as the foundation for devising different computational classes or paradigms within the study of intelligence. To approach the answer to this question, let us examine the different types of intelligences from Section~\ref{sec: intelligences} we just abandoned. What were the motivations for these particular types?

On the surface, a ``dualistic shadow'' produced the distinctions. Physical Intelligence attempted to distinguish between brain and body. But both are physical as there is nothing non-physical. Embodied Intelligence attempted to emphasize the role of the body and its interactions with the brain. Again, this sounds like a dualist distinction. Mechanical Intelligence also attempted to emphasize the role of the body (as opposed to the brain), in particular when viewed from their mechanical properties. But what if the distinction these approaches were attempting to make where not body/brain distinctions but the distinction between different computational paradigms? It is true that the computation performed by the brain and by the body are governed by different physical principles. It is true that mechanics plays a much more prominent role when we want to understand the body. And processes in the brain might be better explained and modeled assuming it performs some kind of digital or analog computation.

What are the different computational paradigms we currently know and how do they differ?  Table~\ref{table: paradigms} lists them. I recommend studying the table for a moment before reading on.

\afterpage{
\begin{landscape}
%\begin{figure}[p]
  \begin{table}
  \footnotesize
 \newcolumntype{Y}{>{\scriptsize\raggedright\arraybackslash}X}
 \begin{tabularx}{23.5cm}{|Y|Y|Y|Y|Y|Y|Y|Y|}
%  \begin{tabularx}{23.5cm}{|X|X|X|X|X|X|X|X|}
  \hline
    \rowcolor[HTML]{C0C0C0}
    \textbf{Paradigm} & \textbf{Digital} & \textbf{Analog} & \textbf{Quantum} & \textbf{Mechanical} & \textbf{Morphological} & \textbf{Neural}\\ % & \textbf{Hybrid} \\ 
    \hline %--------------------------------------------------
    \textbf{Physics} &
    semiconductors, variable electrical conductivity, gating via doping &
    continuous variation aspect of physical phenomena such as electrical, mechanical, or hydraulic quantities) &
    quantum mechanics, entanglement; can be both analog and digital &
    in principle, all of physics, but emphasis has been on rigid structures of certain types (e.g. trusses) &
    mixture of analog and mechanical computing, emphasis on compliant materials and physics phenomena (piezo etc.) &
    biophysics, molecular biology, chemistry \\ % &
%    Combines digital and analog \\
    \hline %--------------------------------------------------
    \textbf{Concept} &
    leverage fast switching of small circuits, ability to compose into complex logic &
    simulate/solve the math problem via physical system with the same constraints/laws &
    exploit parallelism afforded by quantum physics &
    exploit parallelism afforded by mechanics and force propagation within matter (speed of sound) &
    exploit parallelism afforded by mechanics and force propagation within matter (speed of sound) &
    we have not yet decoded how neuronal structures do what they do so well \\ % &
%    ~ \\
    \hline %--------------------------------------------------
    \textbf{Programming principles} &
    symbols, binary, von Neuman, computability, complexity theory, encapsulation &
    signals, calculus, integral, and differentiation, non-deterministic logic &
    too early to tell &
    design principles, smart mechanical design &
    design principles, smart mechanical design, co-design &
    network architecture, error backpropagation, loss functions, data engineering  \\ %&
%    ~  \\
    \hline %--------------------------------------------------
    \textbf{Functional components} &
    transistors, diodes, gates &
    resistors, capacitors, amplifiers, multipliers, potentiometers, function generators &
    qbits (photonics, trapped ions, ...), quantum registers, quantum reversible gates, quantum processing unit &
    materials, joints, mechanical assemblies &
    materials &
    neurons, ion channels  \\ %&
%    ~ \\
    \hline %--------------------------------------------------
    \textbf{Computational capabilities} &
    symbol manipulation, number crunching, storage &
    dynamic and differential systems, complex continuous functions &
    branch and test, search &
    interfacing with the real world while implicitly addressing uncertainty &
    interfacing with the real world while implicitly addressing uncertainty & 
    identifying patterns in data  \\ %&
%    ~ \\
    \hline %--------------------------------------------------
    \textbf{Advantages} &
    generality, ease of programming &
    speed, inherently addresses uncertainty, works on continuous data, high mean time between failure &
    leveraging parallel universes &
    vastly parallel &
    vastly parallel, self-correcting physical interactions, uncertainty suppression and reduction &
    general, can use data, many parameters  \\ %&
%    ~ \\
    \hline %--------------------------------------------------
    \textbf{Disadvantages} &
    limited parallelism, requires explicit representation of uncertainty, memory bottlenecks &
    lack of precision, programming tedious, sensitivity to electric noise &
    not generally usable yet, cooling requirements &
    programming based on vague principles, programs difficult to change &
    programming based on vague principles, programs difficult to change, co-design not yet understood &
    requires much data, many parameters, energy consumption  \\ %&
%    ~ \\
    \hline
\end{tabularx}
\caption{Overview of computational paradigms: The term ``vastly parallel'' refers to the fact that every part of the agent performs computation via atomic forces, for example.}
\label{table: paradigms}
  \end{table}
%  \end{figure}
\end{landscape}
}

It is difficult to explicate the relationship among the computational paradigms listed in Table~\ref{table: paradigms}. Of course, all the way at the bottom, everything is implemented based on physics, probably quantum mechanics. But as a result of quantum mechanics, a number of physical phenomena arise at different levels of abstractions. These phenomena can be leveraged to perform computation, or we might say that these processes are shaped so that they represent a model of the abstract computation we want to perform. For example, analog computation are based on continuous, physical processes of currents and voltages in electric circuits as a means of computation. Underneath it, of course, must lie quantum physics, but it is useful to characterize the resulting higher-level phenomena and to leverage them to encode and perform computation.

Mechanical computation is a specific form of analog computation, emphasizing a particular type of analog phenomenon, i.e., the motion of matter at the macro scale rather that at the electron scale, as it was the case for analog computation based on electricity. For the latter, it would be appropriate to characterize computation based on currents or at the atomic/molecular scale, or on the flow of liquids in the case of hydraulic computations. In contrast, mechanical computation can probably most easily realized by considering kinematics and material properties---phenomena at the super-molecular level. Consequently, different computational paradigms also capture physical phenomena at different levels of abstraction. 

Morphological computation can simply be viewed as a narrower category of mechanical computation but it might differentiate itself from the latter by taking into account the fact that systems based on morphological computation also include some form of digital or neural computation. This necessitates the negotiation of a suitable interface between these two paradigms. Morphological computation might thus be the first computational model explicitly designed for multi-paradigm computation. In contrast to hybrid computation, which combines digital and analog computation across an explicit interface, the morphological computation paradigm not only includes the combination of two paradigms but also reflects the realization that problems must be divided between the paradigms appropriately. This poses a new kind of problem, namely the problem of suitably factorizing the original problems based on the computational strengths of the paradigms involved. 

The key take-away from Table~\ref{table: paradigms} is that each computational paradigm leverages different physics phenomena to perform computation. As a result, the computational capabilities and programming approaches are also different. Each paradigm comes with its unique sets of strengths and weaknesses. An agent leveraging several of these computational paradigms can probably generate a richer set of behaviors, given a finite amount of resources. And the agent can probably generate the same behaviors more efficiently when compared to an agent only leveraging a single paradigm. Intuitively, this seems plausible as a general rule. Practically, there are many specific examples (all species on planet earth). 

On the other hand, in some formal sense, these paradigms possess probably equivalent computational power with respect to the Church-Turing hypothesis~\citep{Hopcroft-1979}. They all can, in principle, compute all computable functions. But these arguments from the domain of computability are not our focus when we discuss computation to generate intelligent behavior. Already very simple problems relevant to intelligence, for example in motion planning of polygons among polygonal obstacles, are NP-complete~\citep{latombe-91} and therefore computationally very challenging. Nevertheless, intelligent agents solve versions of these problems all the time. Therefore, we need to shift our attention away from completeness towards bounded-resource, practical implementations of computational solutions. To judge a problem, it will not be helpful to ask how much computation in the different paradigms is required to solve \textit{all} possible instances of that problem. An important question will be if there are large and relevant subsets of all possible problems that can be solved with reasonable efficiency---within one of the computational paradigms. An example of this is factorization of large integers, which becomes highly efficient using quantum computation but is practically intractable with digital computation.

As a result, building an intelligent agent becomes easier if we pick the computational paradigm most suitable for expressing a necessary computation. When we look at the computational paradigms listed in Table~\ref{table: paradigms}, we should foremost think about practical feasibility: Does the inclusion of a computational paradigm permit the construction of an agent with novel capabilities? These additional capabilities will most likely result from our ability to express a solution in that paradigm more simply, rather than from the inherent computational capabilities associated with the paradigm (as we assume these are the same for all paradigms, at least in principle). Effectively, we consider whether the inclusion of a paradigm can reduce the overall cost, be it of constructing the agent or of performing the computation.

I would like to suggest that the computational paradigms described here are a useful way of specific structuring subsets of physical processes. Each paradigm captures a particular exploitable regularity of physics, forming such a subset. Associated with each paradigm is also a suitable abstraction, allowing us to reason about the particular type of physical process (or computation).

The computational paradigms let us unite the different physical processes going on in agents into a single multi-paradigm computation. For example, a biological agent performs neural computations in the brain to determine its behavior. But these computations alone do not determine the behavior in its entirety. In addition, just to give an example, the lens and shape of the eye determine how the environment is perceived. And the structure of the skeletal and muscular structure lets neural instructions play out differently each time, depending on which forceful interactions with the environment occur during the behavior. Therefore, behavior is not determined by neural computations alone, mechanical computations also affect it. The resulting behavior is multi-paradigmatic. 

Given the classification of physical processes into computational paradigms, and given the view that these computational paradigms are composed in the generation of intelligent behavior, we must then understand how to best perform such compositions to understand and synthesize intelligence. This requires knowledge of which computational paradigm is suited best for a particular category of problems or behavior. It will become important to understand this paradigm/problem relationship at a general level. In the context of digital computation, this relationship is already well understood via complexity and language theory~\citep{Hopcroft-1979}. We know that finite automata, push-down automata, and Turing machines implicitly capture regular languages, context-free languages, and recursively enumerable languages, respectively~\citep{Hopcroft-1979}, just to give an example. This is a very helpful understanding of the relationship between problems (regular languages) and the most adequate computational solution (finite automata). I am not aware of similar specific results for other computational paradigms. And maybe the success of digital computation is linked to the ease with which the relationship between computation and problem can be established. But it seems highly desirable to attempt to characterize the computations that can be performed within each of the computational paradigms.

%----------------------------------------------------------------------
\subsection{The Impact of Computational Paradigms on the Notion of Computation}
%----------------------------------------------------------------------

The question of when a physical system performs computation and when not remains debated~\cite{Horsman-2013,sep-computation-physicalsystems}. The view proposed here deviates from some of the positions taken in the literature and thereby seems to resolve some of the difficulties in this discussion. I would like to briefly mention those, even though the notion of computation is not the main focus of this text.

\begin{enumerate}
\item
  One criticism of pancomputationalism is that the notion of
  computation becomes vacuous when \textit{everything} is considered
  computation~\cite{Horsman-2013}. The view propagated here is that
  every physical process \textit{can} be a computation if it
  contributes to the behavior of an agent. Therefore, not
  \textit{everything} is considered computation, the set of physical
  processes is subdivided into a meaningful way.
\item
  The set of physical processes considered as computation is further
  reduced by the requirement that physical processes are categorized
  into computational paradigms. Each paradigm has its own abstraction
  that we can use to model, understand, and program. By selecting a
  subset of all paradigms to create an intelligent agent, the
  criticism of vacuousness~\cite{Horsman-2013} is removed.
\item
  Central to many arguments about computation is the
  \textit{representation relation} from physics, linking physical
  processes to abstract mathematical objects. However, I believe that
  this framework, while very useful for some kind of arguments, makes
  restrictive assumptions. This can be seen, for example, when the
  mathematical object becomes a necessity for computation and is
  required to be known, even a priori~\cite{Horsman-2013}: ``If a
  computational description of a physical evolution can only be
  applied post-hoc, then the system has not acted as a computer.'' But
  in our view, a computational description (mathematical object) is
  not required at all. The result of a computation does not need to
  represent a symbol or mathematical object, and it does not need to be
  predicted to be useful for the behavior of an agent. It seems that
  with the \textit{representation relation} one has inadvertently
  imported unnecessary constraints.
\item
  Computational paradigms resolve a confusion about embodiment
  prevalent in the discussion. Take a statement like the following:
  ``The centrifugal governor is not a computer, even if it can be
  replaced by a computational system that measures the relevant
  values, computes a response, and controls the relevant value (valve
  opening)''~\cite{Muller-2017}. It does not seem plausible that the
  same function realized in two different physical processes should be
  called computation in one case and not in another. Both the
  centrifugal governor and the alternative computational system are
  simply matter arranged in such a way that the behavior is
  achieved. We might be led into believing that they are fundamentally
  different because they implement different computational
  paradigms. In the view proposed here, they both perform computation
  and result in the same behavior.
\item
  Another discussion in the literature pertains to systems that
  combine multiple computational paradigms. Authors object to the
  notion of ``offloading'' or ``outsourcing''
  computation~\cite{Muller-2017}, terms often used in the context of
  morphological computation~(see Section~\ref{morph comp}), when
  mechanical computation enables the simplification of digital
  computation~\cite{Bhatt-21}. This situation is easily explained in
  the context of computational paradigms. Let us use soft manipulation
  as an example~\cite{Bhatt-21}. A soft-material hand contributes
  importantly to the manipulation behavior through its carefully
  designed compliance, enabling complex in-hand manipulation with
  nearly no digital computation~\cite{Bhatt-21}. In such a scenario,
  we can say that the overall behavior of in-hand manipulation is
  produced by combining two computations, each from a different
  computational paradigm (mechanical and digital computation). Because
  the responsibility for producing the behavior is distributed aptly
  across these two types of computation, the resulting behavior is
  highly robust and simple. The question of whether computation is
  outsourced or offloaded is replaced by the distribution of
  responsibility for behavior generation across two computational
  paradigms. The complications previously perceived~\cite{Muller-2017}
  have disappeared.
\end{enumerate}

%======================================================================
\section{Intelligences---Status Quo Revisited}
\label{sec: intelligences revisted}
%======================================================================

Let us reexamine the definitions of intelligences from Section~\ref{sec: intelligences} based on the proposed multi-paradigmatic computational structure. We will see that computational paradigms resolve the difficulties we discovered earlier.  For example, Physical Intelligence emphasizes analog computation, mechanical computation, and morphological computation (not the body, as stated in the definition). And it contrasts this with digital computation and neural computation (not the brain, as stated in the definition). Similarly, Embodied Intelligence sought to address the interplay between brain and body, but really it is about how to best divide a particular problem across digital and analog/mechanical computation and how to have those two computations integrate well. 

In Section~\ref{sec: intelligences}, we also discussed the term ``embodied'' as a property of intelligence. Let us contextualize this term in light of computational paradigms. For example: does being embodied imply the inclusion of mechanical computation?  CPUs are a counter-example, where digital computation is embodied without mechanical computation. Even a robot does not necessarily have to include (significant) mechanical computation. If a robot's body is very stiff, controlled at high frequencies, and with very high torque, then the robot's behavior is entirely determined by the digital control program, without mechanical computation. Mechanical computation enters the picture once the robot engages in physical interactions with the compliant environment. Therefore, the term ``embodiment'' does not help to differentiate between computational paradigms. And as a result---if we assume that computational paradigms are important to our discussion---the term ``embodied'' is not very helpful in intelligence research.

%======================================================================
\section{Intelligence as Computation}
\label{sec: intelligence as computation}
%======================================================================

We are now ready for moving towards a description of the nature of intelligence.
\textit{Intelligence is the set of computations that satisfy certain properties.} What exactly these properties are will have to be the subject of intelligence research but I will make some suggestions below. To implement these computations we have various computational paradigms at our disposal. As our understanding of these paradigms improves, the task of producing intelligent computation will become easier. 

This proposed view of intelligence---it cannot be called a definition because it is still missing important information---as a subset of all computations has advantages over current attempts to define intelligence~\citep{CambridgeHandbookOfIntelligence-2011}. Current attempts define intelligence by enumerating characteristics of intelligent behavior. Instead, I am proposing to characterize the computational processes that produce the behavior. This promises to be much simpler. The characteristics of behavior are the result of the mechanism and its interactions with diverse environment. It seems plausible that behavior is much more complex and versatile than the mechanism that generates it. It should therefore be simpler to characterize the computational mechanism.

The proposed view of intelligence is not subject to anthropomorphic bias. Intelligence can occur in individuals, in collectives, in societies, but also at behavioral, cellular, or genetic levels. Viewing intelligence as computation opens the mind to the discovery of intelligence that differs from human intelligence. I therefore believe that the proposed view of intelligence as computation provides a productive conceptual foundation for intelligence research.

Let us enumerate the key questions that our field must answer, if we assume this view of intelligence as computation:

\begin{enumerate}
\item Understand better existing computational paradigms, including what types of problems they are particularly suited for
\item Develop an engineering practice for designing and programming systems that combine several computational paradigms
\item Identify the specific properties of computation that characterize intelligent computation, and, if such a characterization has been postulated,
\item Learn how to leverage these properties as inductive biases for producing synthetic intelligent agents and for understanding natural intelligent agents
\end{enumerate}

It is important to note that the answer to these questions will most likely vary significantly for different behavioral spaces, such as the one spanned by individuals, collectives, societies, cells, genes, just to name the same examples as above.

Let us speculate about the properties of computation that produce intelligence. These properties can serve as inductive biases for building intelligent agents or for explaining intelligent behavior. When designing and programming an agent, these properties facilitate the generation of intelligent computation. I would speculate that these properties will turn into programming principles. The more of these principles we accumulate and are able to apply, the easier will it become to build intelligent agents and the more intelligent the resulting agents can become.

It remains possible that intelligence will never be characterized as a sharply defined set of computations. We might continue to discover new properties or principles, specific to particular ecological niches, without ever reaching consensus on a fixed set of properties. It is possible that different intelligent agents are based on some properties only, or even different properties that produce the same behavior. There might be properties that are necessary and other that are sufficient. So we might have to resort to the fall-back statement that intelligence is just a form of complex computation. But to me it seems exciting to attempt to find a succinct description of what distinguishes intelligence from other computations, in particular as it promises to enable the construction of intelligent agents.

The following subsections present speculations about possible properties that might help discriminate intelligent computation. If they turn out to be necessary principles for exhibiting intelligence, we could call them principles of intelligent computation, or even the principles of intelligence. This would provide support to the view that intelligence is computation that satisfies certain properties of sophistication, characterized by these principles.

%----------------------------------------------------------------------
\subsection{Multiple Computational Paradigms}
%----------------------------------------------------------------------

It might be an inherent property of intelligent computation that it leverages multiple computational paradigms. This is certainly true for all forms of biological intelligence, which leverage neural computation and mechanical computation.

It is also possible that the path towards determining a minimal set of properties to characterize intelligence or the path towards a form of intelligence that is only based on a single computational paradigm will require us to discover many properties and to exploit many computational paradigms. If we were to understand intelligence completely, a small set of properties might suffice to characterize intelligent computation, and a single computational paradigm might suffice to implement it.

%----------------------------------------------------------------------
\subsection{Diverse, Active Component Interactions}
\label{sec: diverse interactions}
%----------------------------------------------------------------------

The intrinsic complexity of the computational program executed by an intelligent agent, possibly measured by something like Kolmogorov complexity~\citep{kolmogorov1965three}, is substantially less than the complexity of the behavior it is capable of producing. This discrepancy results from the fact that the same computational agent can interact with a very large number of different environments. Every relevant change in the environment can lead to significant changes in behavior. This modification of behavior does not require a fundamental change in the agent's computational structure. Therefore, the goal of the agent remains achievable under large environmental variations without significant changes to the agent. In fact, this seems to be a prerequisite for intelligence, namely, that agents achieve their goals robustly under significant variations in environmental conditions.

But what are the computational properties implemented in the agent that support this kind of robust adaptability to environmental conditions?

I would like to propose the principle of \textit{diverse, active component interaction} as one possible property that contributes to the robustness of intelligent behavior. It is best explained in contrast to current software engineering practice of developing systems with strong modularity, i.e., systems in which the information exchanged between components is specified a priori and designed to be as simple as possible (this results from the software principle of encapsulation)~\citep{martin-2024}. The advantage of the software engineering approach is that errors and failure modes are confined to specific components of the overall agent. However, this also leads to brittleness under environmental changes. If the system is confronted with a novel environmental situation, all components might be able to contribute to a useful behavioral adaptation, even outside of the initially constrained, pre-programmed modularization of the system. For sure, any useful adaptation to new circumstances must be coordinated among several of the computational components of the agent. This coordination cannot easily be anticipated a priori, otherwise the agent would have to be as complex as the behavior it produces (and software engineers would have to be able to anticipate all possible environmental variations). Therefore, the information exchange between components must be \textit{active} to adjust to the circumstances. It must also be \textit{diverse} so that in many situations the required information can be extracted and exchanged. 

Using \textit{diverse, active component interaction} as a principle of computation, we can hope to create systems that achieve their programmed goals robustly, even when confronted with environmental variations. The principle implies that different components of the computational systems are capable of engaging with other components in diverse interactions, facilitated by active and diverse information channels. The components themselves must be able to change their behavior in response to the information relayed to it.

\textit{Diverse, active component interaction} is, in many ways, the opposite of good software engineering practice, where the interface between component is fully specified and often as simple as possible to support encapsulation and high modularity~\citep{martin-2024}. However, if variations in behavior are required by new environmental circumstances, it is only plausible that these variations must be brought about by different responses of the system's components. These responses must be coordinated with other components. Also, components might facilitate the adequate response of a different component by providing information to it that would otherwise remain unknown. Therefore, having diverse component interactions might be an important ingredient to support this behavioral diversity on the way to building robust intelligent systems.

What I have stated about this principle is far from being a concrete recipe for implementing robust systems. But employing a vague understanding of this principles in system building has already led my group to building very robust and award-winning robotic systems~\citep{Eppner-2018}. There also is fascinating evidence in biology and medicine, demonstrating that at least some specific biological components  are set up to support strong component interactivity at the genetic~\citep{leff_complex_1986}, cellular~\citep{vaney_gap_2000}, and tissue levels~\citep{blackiston_ectopic_2013}.

At the genetic level, there is a process called alternative splicing~\citep{leff_complex_1986} that exhibits strong component interactivity. Alternative splicing is part of the production process of proteins inside cells. Genes encode the blueprint of proteins. However, proteins are not always encoded in a single, dedicated gene. Instead, alternative splicing assembles several pieces of genetic information to produce the protein. These pieces are reused in different combinations to produce different proteins. This implies that this genetic information must be ready for being combined with different pieces of information, each time producing a protein vital for the survival of the cell. The individual pieces do not only contain the necessary information to produce a single protein, they can be combined in different ways to produce different proteins.

Another example: Cells are connected by so-called gap-junctions~\citep{vaney_gap_2000} to facilitate strong component interconnectivity among cells. These junctions enable flexible communication across cell membranes, with the set of possible signal transmissions referred to as connexin. This enables flexible cellular communication adaptable to a broad range of environmental conditions. The communication, in turn, enables different modes of interaction between cells.

A third example, at the tissue level, is even more surprising. Surgically attaching a tadpole eye to the tail of another tadpole, triggers responses in the tissue that enables the tadpole to integrate the information obtained from that ``Frankenstein eye''~\citep{blackiston_ectopic_2013}. The tissue of the tadpole responds flexibly to the unexpected sensor information, growing a connection between the optical nerve and the tadpole's nervous system. The tadpole begins to change its behavior in response to light perceived by this new eye. This demonstrates that the tissue of tadpoles is capable of engaging in new, evolutionarily unplanned component interactivity.

Another example of diverse, active component interactions comes form transplantation medicine. If intestines are transplanted, the inclusion of the liver in the transplant appears to reduce the probability of rejection of the implanted organs~\citep{vianna_liver_2023}. This is an example of a general pattern~\citep{fridell_survival_2019}, in which the inclusion of a weakly immunigenic organ can aid the long-term prospects for the transplant of a more strongly immunigenic organ. This, too, demonstrates that there are active interactions among organs that go beyond their immediate bodily function.

These examples illustrate that biological agents are capable of strong component interactivity at different levels of abstraction. It is plausible that this principle is a critical enabler of the robustness exhibited by biological systems. And it is equally plausible that this architectural pattern can also be employed to large benefits at higher levels of abstraction. 

Diverse, active component interactions can only be realized if components are developed via co-design, as the flexible coordination of components now becomes an important requirement for all components. To build this into components, co-design, i.e., the joint development of multiple components seems the only plausible way. In addition, we must realize that ``physical interaction with the environment as the primary source of constraint on the design of intelligent systems''~\citep{brooks_elephants_1990}. Therefore, co-design must occur in realistic settings, possibly in rich real-world ecological niches, to produce relevant component interactions that in turn produce robust adaptation of behavior to environmental variations.

%----------------------------------------------------------------------
\subsection{Agent/Environment Computation}
%----------------------------------------------------------------------

The principle of \textit{diverse, active component interaction}, discussed in the previous section also applies to the components ``agent'' and ``environment''. The computation that constitutes intelligence happens on both sides of this boundary~\citep{clark_supersizing_2010,Murphy-Paul-2022}. Diverse interactions between both components are critical to enable robust and intelligent behavior. This is closely related to the \textit{embodiment hypothesis}~\citep{smith_development_2005}, which states ``that intelligence emerges in the interaction of an agent with an environment and as a result of sensorimotor activity.''

Given the discussion above, the attributive noun ``embodiment'' might be incorrect. The key property one needs to emphasize is not that the agent is embodied (all agents are), it is that the computation that takes place is distributed between agent and environment. This fact might be another candidate for a principle of intelligence. The claim would be that the generation of intelligent behavior requires or is greatly facilitated by incorporating aspects of the environment into the computation. There is ample support from cognitive science in support of this happening in humans~\citep{triesch_what_2003}.

The notion of ``sensorimotor activity'' describes the diverse interactions that occur across the agent/environment boundary. Our own experience confirms that both our perceptual abilities and our abilities for action are diverse and can support a broad range of robust and intelligent behaviors, i.e., they enable diverse component interactions across the agent/environment boundary. For this, it seems critical that the same computational paradigm, namely mechanical computation, is available on either side of the interaction. This might be the real reason, rather than embodiment, as a form of embodiment could be limited to a different computational paradigm. Therefore, it might be more accurate to call the embodiment hypothesis the ``identical computational paradigm across the boundary'' hypothesis. This hypothesis might state that if the same computational paradigm is available on both sides of the agent/environment boundary, implementation complexity goes down, computational complexity goes down, and robustness and versatility go up.

%----------------------------------------------------------------------
\subsection{More On Hypothesized Principles}
%----------------------------------------------------------------------

As a reminder: my proposal for understanding intelligence from a computational perspective is to identify principles that characterize intelligence. This could enable intelligent systems, realized by appropriately implementing these principles in a computational system. Of course, identifying these necessary principles requires a long-term, community-wide research effort. This effort will require the close collaboration of researchers studying natural and artificial intelligence. Above, I discussed three possible principles, but there must be many more, of course.

A simple candidate for a principle of computational intelligence is the ``similarity principle''~\citep{chang_2017}. It states that similar things behave similarly---or perform similar computation. But since this applies to nearly all physical processes at the macro-level, we cannot use it to discriminate general computation from intelligent computation.

There are likely many more principles we must identify before we can hope to fully characterize intelligence or easily create intelligent artificial systems. For example, intelligent systems are likely to perform computations at many different levels of abstractions. That could be one of the necessary principles. And across these levels, similar computational principles might apply, giving rise to a new principle: self-similarity across levels of abstractions. In Section~\ref{sec: diverse interactions} we saw that in biological systems the principle of \textit{diverse, active component interactions} occurs at the genetic, cellular, and organ levels. This is an example of the self-similarity principle in biology. The self-similarity principle might also apply to representations across these levels. It will be an interesting scientific journey to hypothesize such principles and to validate them in biological and artificial agents.

%======================================================================
\section{Cognition Versus Intelligence}
\label{sec: cognition versus intelligence}
%======================================================================

Cognition and Intelligence are closely related concepts. However, I would say that no operationalizable and agreed-upon definition of either term exists. This makes it difficult to articulate their difference exactly. And somehow cognitive science and intelligence research have co-existed productively without this clarification. 

Still, the computational view of intelligence implies a useful relationship between cognition and intelligence. Cognition, I would like to suggest, is concerned with the mechanism of computation that produce intelligence: What algorithms are implemented? What representations are required? How do the algorithms perform and what are their inherent limitations? Just to name a few. In contrast, intelligence refers to how these computations produce behavior that solves problems and achieves goals. 

Let us exploit the computational metaphor to the fullest and transfer these definitions of cognition and intelligence to the world of digital computers. Cognition would be the algorithms representations implemented in the computers hardware (binary representation of numbers and, for example, the Dadda tree as a hardware implementation for multiplication) as well as the functionality implemented in the BIOS,  operating system, and applications. Intelligence would correspond to the interactions between computer and user (the environment) to solve certain problems. We would like to think that this analogy between a user and the environment is weak because the environment does not ``drive'' the interaction the same way the user does, but this point is worth contemplating.

In some sense, this view resembles Nikolaas Tinbergen's four questions~\citep{tinbergen_aims_1963}. Tinbergen, a Nobel Prize laureate and founder of behavioral biology, postulated that to fully understand the behavior of a biological agent, one must explain the causation, ontogeny, evolution, and survival value of the observed behavior. Without going into details, Tinbergen argued that answers given from four different perspectives are necessary to obtain the full picture. I would like to suggest that cognition and intelligence represent two perspectives from which we can begin to explain intelligent behavior. There might be more (Tinbergen's four questions are very promising candidates) and each perspective must be spelled out in more detail, but maybe this paper can serve as a starting point for this debate within our community.

%======================================================================
\section{Consequences for Embodiment and Flavors of Intelligence}
%======================================================================

Understandably, after decades of focus on digital computation, a counter-movement is now emphasizing the importance of analog computation and mechanical computation. This has led to branches of intelligence/cognition research named Embodied Intelligence or Embodied Cognition. The term ``body'' was intended to emphasize the non-brain portion of the body, the portion of the body that does not perform neural computation but rather mechanical computation. But this led to the issues we already discussed above: All of the body, including the brain, contributes to the computation performed by a biological agent. The emphasis of ``body'' does not disambiguate and does not emphasize the right concept. Speaking of neural and mechanical computation does accomplish the desired distinction.

One of the original accounts of \textit{embodiment} put it like this~\citep{varela_embodied_1993}:
\begin{quotation}
  By using the term embodied we mean to highlight two points: first that cognition depends upon the kinds of experience that come from having a body with various sensorimotor capacities, and second, that these individual sensorimotor capacities are themselves embedded in a more encompassing biological, psychological and cultural context.
  \end{quotation}

And this rings true today as much as it did in 1993. A detailed discussion of the evolution of embodiment from a philosophical perspective has been provided by~\citet{sep-embodied-cognition}. In my view, all accounts of embodied cognition or embodied intelligence fall short of fully articulating what exactly the contribution is that is made by the body. My answer, as outlined above, is: mechanical computation, as contrasted by neural computation performed in the nervous system. What is falsely referred to as \textit{embodiment}---because everything is embodied---actually refers to the addition of an additional computational paradigm, i.e., mechanical computation.

If the reader buys into the arguments laid out above, then we should not talk about embodiment, embodied cognition, or embodied intelligence any longer. Instead, we should talk about intelligence based on multiple computational paradigms. In the case of biological agents, we know that they rely at least on neural computation and mechanical computation. We can unite the different flavors in a study of intelligence that can clearly articulate the type of computation employed by an agent and what computational power it contributes to the behavior.

%======================================================================
\section{Conclusion}
%======================================================================

Intelligence is a computational phenomenon. Computation comes in various flavors, including digital computation, analog computation, quantum computation, mechanical computation etc. When researching intelligence and when talking about it, we should refer to these different computational paradigms to explain how an observed behavior comes about. This is much cleaner and more conducive to progress than the ongoing debates about different flavors of intelligences. Unfortunately, these flavors are now associated with different communities and academic careers. This will make it difficult to leave these distinctions behind. But I think the reward for doing so would be substantial. We should realize that, given the diversity in computational phenomena contributing to intelligence, only a unified research program can move towards a constructive understanding of intelligence. All computational paradigms can contribute to intelligence, but, more importantly, their interactions are poised to play an equally important role. How can we best combine digital computation, maybe in the form of neural computation, and mechanical computation? This question can only be answered by studying these computational paradigms together. This, of course, is not a call for dismantling disciplinary research programs---not at all. They are important and will continue to be important. It is a call for breaking down the barriers that are keeping us from making progress towards a unified science of intelligence.

%======================================================================
\section*{Acknowledgments}
%======================================================================

I am thankful for interesting discussions and helpful comments from Alan Akbik, Aravind Battaje, Guillermo Gallego, Olaf Hellwich, Alex Kacelnik, Matt Mason, Vito Mengers, Christa Th\"one-Reineke, and Oscar Wei{\ss}enbach. This work was funded by the Deutsche Forschungsgemeinschaft (DFG, German Research Foundation) under Germany's Excellence Strategy  EXC 2002/1 ``Science of Intelligence'' project number 390523135. 

%======================================================================
%\section*{References}
%======================================================================
\newcommand{\newblock}{}
\bibliographystyle{abbrvnat}
\balance
\bibliography{references}

\end{document}